\documentclass[11pt,a4paper]{article}
\usepackage[hyperref]{naaclhlt2018}
\usepackage[utf8]{inputenc}
\usepackage{times}
\usepackage{graphicx}
\usepackage{latexsym}
\usepackage{amssymb}
\usepackage{amsmath,amsthm}
\usepackage{bm}
\usepackage{booktabs}
\usepackage{hyperref}
\usepackage{url}

\newcommand{\source}{\mathbf{x}}
\newcommand{\sourcetok}{x}

\newcommand{\pseudoreference}{\mathbf{u}^*}

\newcommand{\goldreference}{\mathbf{t}}
\newcommand{\goldreferencetok}{t}

\newcommand{\candhypos}{\mathcal{U}}
\newcommand{\candhypo}{\mathbf{u}}
\newcommand{\candhypotok}{u}

\newcommand{\TokNLL}{\texttt{TokNLL}}
\newcommand{\TokLS}{\texttt{TokLS}}
\newcommand{\SeqNLL}{\texttt{SeqNLL}}
\newcommand{\Risk}{\texttt{Risk}}
\newcommand{\MaxMargin}{\texttt{MaxMargin}}
\newcommand{\MultiMargin}{\texttt{MultiMargin}}
\newcommand{\SoftmaxMargin}{\texttt{SoftmaxMargin}}
\newcommand{\Constrained}{\texttt{Constrained}}
\newcommand{\Weighted}{\texttt{Weighted}}

\aclfinalcopy 

\DeclareMathOperator*{\argmax}{arg\,max}

\title{Classical Structured Prediction Losses for Sequence to Sequence Learning}

\newcommand*\samethanks[1][\value{footnote}]{\footnotemark[#1]}
\author{
  Sergey Edunov\thanks{\hspace{0.075in}Equal contribution.} , Myle Ott\samethanks\hspace{0.05in}, \\
  {\bf Michael Auli, David Grangier, Marc'Aurelio Ranzato} \\
  Facebook AI Research \\
  Menlo Park, CA \and New York, NY
}

\date{}

\begin{document}

\maketitle

\begin{abstract}
There has been much recent work on training neural attention models at the sequence-level using either reinforcement learning-style methods or by optimizing the beam. In this paper, we survey a range of classical objective functions that have been widely used to train linear models for structured prediction and apply them to neural sequence to sequence models.
Our experiments show that these losses can perform surprisingly well by slightly outperforming beam search optimization in a like for like setup.
We also report new state of the art results on both IWSLT'14 German-English translation as well as Gigaword abstractive summarization. 
On the large WMT'14 English-French task, sequence-level training achieves 41.5 BLEU which is on par with the state of the art.\footnote{An implementation of the losses is available as part of fairseq at:\\
\url{https://github.com/pytorch/fairseq/tree/classic_seqlevel}}
\end{abstract}

\section{Introduction}

Sequence to sequence models are usually trained with a simple token-level likelihood loss \citep{sutskever2014sequence,bahdanau2014neural}.
However, at test time, these models do not produce a single token but a whole sequence. In order to resolve this inconsistency and to potentially improve generation, recent work has focused on training these models at the sequence-level,
for instance using REINFORCE \citep{ranzato2015sequence}, actor-critic \citep{bahdanau2016ac}, or with beam search optimization \citep{wiseman2016acl}.

Before the recent work on sequence level training for neural networks, there has been a large body of research on training linear models at the sequence level.
For example, direct loss optimization has been popularized in machine translation with the Minimum Error Rate Training algorithm (MERT; Och 2003)\nocite{och:2003:acl} and expected risk minimization has an extensive history in NLP~\cite{smith+eisner2006acl,rosti:2010:wmt,green:2014:wmt}.
This paper revisits several objective functions that have been commonly used for structured prediction tasks in NLP \citep{gimpel+smith2010acl} and apply them to a neural sequence to sequence model \citep{gehring2017icml} (\textsection\ref{sec:model}).
Specifically, we consider likelihood training at the sequence-level, a margin loss as well as expected risk training.
We also investigate several combinations of global losses with token-level likelihood.
This is, to our knowledge, the most comprehensive comparison of structured losses in the context of neural sequence to sequence models (\textsection\ref{sec:objectives}).

We experiment on the IWSLT'14 German-English translation task~\citep{cettolo2014report} as well as the Gigaword abstractive summarization task~\citep{rush2015abs}.
We achieve the best reported accuracy to date on both tasks.
We find that the sequence level losses we survey perform similarly to one another and outperform beam search optimization~\citep{wiseman2016acl} on a comparable setup.
On WMT'14 English-French, we also illustrate the effectiveness of risk minimization on a larger translation task.
Classical losses for structured prediction are still very competitive and effective for neural models (\textsection\ref{sec:expsetup}, \textsection\ref{sec:results}).

\section{Sequence to Sequence Learning}
\label{sec:model}

The general architecture of our sequence to sequence models follows the encoder-decoder approach with soft attention first introduced in \cite{bahdanau2014neural}.
As a main difference, in most of our experiments we parameterize the encoder and the decoder as convolutional neural networks instead of recurrent networks \citep{gehring2016convolutional,gehring2017icml}. Our use of convolution is motivated by computational and accuracy considerations. However, the objective functions we present are model agnostic and equally applicable to recurrent and convolutional models. We demonstrate the applicability of our objective functions to recurrent models (LSTM) in our comparison to \citet{wiseman2016acl} in \textsection\ref{sec:bso}.

\smallskip
\noindent {\bf Notation.} We denote the source sentence as $\source$, an output sentence of our model as $\candhypo$, and the reference or \emph{target} sentence as $\goldreference$.
For some objectives, we choose a pseudo reference $\pseudoreference$ instead, such as a model output with the highest BLEU or ROUGE score among a set of candidate outputs, $\candhypos$, generated by our model.

Concretely, the encoder processes a source sentence $\source = \left( \sourcetok_1, \dots, \sourcetok_m \right)$ containing $m$ words and outputs a sequence of states $\mathbf{z} = \left( z_1. \dots, z_m \right)$.
The decoder takes $\mathbf{z}$ and generates the output sequence $\candhypo = \left( \candhypotok_1, \dots, \candhypotok_n \right)$ left to right, one element at a time.
For each output $\candhypotok_i$, the decoder computes hidden state $h_i$ based on the previous state $h_{i-1}$, an embedding $g_{i-1}$ of the previous target language word $\candhypotok_{i-1}$, as well as a conditional input $c_i$ derived from the encoder output $\mathbf{z}$.
The attention context $c_i$ is computed as a weighted sum of $\left( z_1, \dots, z_m \right)$ at each time step.
The weights of this sum are referred to as attention scores and allow the network to focus 
on the most relevant parts of the input at each generation step.
Attention scores are computed by comparing each encoder state $z_j$ to a combination of the previous decoder state $h_i$ and the last prediction $\candhypotok_i$;
the result is normalized to be a distribution over input elements.
At each generation step, the model scores for the $V$ possible next target words $\candhypotok_i$ by transforming the decoder output $h_i$ via a linear layer with weights $W_o$ and bias $b_o$: $s_i = W_o h_i + b_o$.
This is turned into a distribution via a softmax:
$p(\candhypotok_i | \candhypotok_1, \dots, \candhypotok_{i-1}, \source) = \operatorname{softmax}(s_i)$.

Our encoder and decoder use gated convolutional neural networks which enable fast and accurate generation \cite{gehring2017icml}. Fast generation is essential to efficiently train on the model output as is done in this work as sequence-level losses require generating at training time.
Both encoder and decoder networks share a simple block structure that computes intermediate states based on a fixed number of input tokens and we stack several blocks on top of each other.
Each block contains a 1-D convolution that takes as input $k$ feature vectors and outputs another vector; subsequent layers operate over the $k$ output elements of the previous layer. The output of the convolution is then fed into a gated linear unit \citep{dauphin2017icml}.
In the decoder network, we rely on causal convolution which rely only on states from the previous time steps. The parameters $\mathbf{\theta}$ of our model are all the weight matrices in the encoder and decoder networks. Further details can be found in \citet{gehring2017icml}.

\section{Objective Functions}
\label{sec:objectives}

We compare several objective functions for training the model architecture described in \textsection\ref{sec:model}. The corresponding loss functions are either computed over individual tokens (\textsection\ref{sec:tokobj}), over entire sequences (\textsection\ref{sec:seqobj}) or over a combination of tokens and sequences (\textsection\ref{sec:comb}). An overview of these loss functions is given in Figure~\ref{fig:objectives}.

\begin{figure*}
\begin{align}
\label{eq:toknll}
\mathcal{L}_{\TokNLL} = &
-\sum_{i=1}^n \log p(\goldreferencetok_i | \goldreferencetok_1, \dots, \goldreferencetok_{i-1}, \source) \\
\label{eq:tokls}
\mathcal{L}_{\TokLS} = &
-\sum_{i=1}^n \log p(\goldreferencetok_i | \goldreferencetok_1, \dots, \goldreferencetok_{i-1}, \source) - D_{KL}(f\Arrowvert p(\goldreferencetok_i | \goldreferencetok_1, \dots, \goldreferencetok_{i-1}, \source))\\
\label{eq:seqnll}
\mathcal{L}_{\SeqNLL} = &
-\log p(\pseudoreference|\source) + \log \sum_{\candhypo \in \candhypos(\source)} p(\candhypo|\source) \\
\label{eq:risk}
\mathcal{L}_{\Risk} = &
\sum_{\candhypo \in \candhypos(\source)} \operatorname{cost}(\goldreference, \candhypo) \frac{p(\candhypo|\source)}{\sum_{\candhypo' \in \candhypos(\source)} p(\candhypo'|\source)} \\
\label{eq:maxmargin}
\mathcal{L}_{\MaxMargin} = &
\max \left[0, \operatorname{cost}(\goldreference, \hat{\candhypo}) - \operatorname{cost}(\goldreference, \pseudoreference) - s(\pseudoreference | \source) + s(\hat{\candhypo} | \source) \right] \\
\label{eq:multimargin}
\mathcal{L}_{\MultiMargin} = &
\sum_{\candhypo \in \candhypos(\source)}
\max \left[0, \operatorname{cost}(\goldreference, \candhypo) - \operatorname{cost}(\goldreference, \pseudoreference) - s(\pseudoreference | \source) + s(\candhypo | \source) \right] \\
\label{eq:smm}
\mathcal{L}_{\SoftmaxMargin} = &
-\log p(\pseudoreference | \source) + \log\sum_{\candhypo \in \candhypos(\source)} \exp\left[ s(\candhypo|\source) + \operatorname{cost}(\goldreference, \candhypo) \right]
\end{align}
\caption{Token and sequence negative log-likelihood (Equations~\ref{eq:toknll} and~\ref{eq:seqnll}), token-level label smoothing (Equation~\ref{eq:tokls}), expected risk (Equation~\ref{eq:risk}), max-margin (Equation~\ref{eq:maxmargin}), multi-margin (Equation~\ref{eq:multimargin}), softmax-margin (Equation~\ref{eq:smm}).
We denote the source as $\source$, the reference target as $\goldreference$, the set of candidate outputs as $\candhypos$ and the best candidate (pseudo reference) as $\pseudoreference$.
For max-margin we denote the candidate with the highest model score as $\hat{\candhypo}$.
\label{fig:objectives}}
\end{figure*}

\subsection{Token-Level Objectives}
\label{sec:tokobj}

Most prior work on sequence to sequence learning has focused on optimizing token-level loss functions, i.e., functions for which the loss is computed additively over individual tokens.

\subsubsection*{Token Negative Log Likelihood (\TokNLL)}
\label{sec:toknll}
Token-level likelihood (\TokNLL, Equation~\ref{eq:toknll}) minimizes the negative log likelihood of individual reference tokens $\goldreference = \left( \goldreferencetok_1, \dots, \goldreferencetok_n \right)$.
It is the most common loss function optimized in related work and serves as a baseline for our comparison.

\subsubsection*{Token NLL with Label Smoothing (\TokLS)}
\label{sec:tokls}
Likelihood training forces the model to make extreme zero or one predictions to distinguish between the ground truth and alternatives.
This may result in a model that is too confident in its training predictions, which may hurt its generalization performance.
Label smoothing addresses this by acting as a regularizer that makes the model less confident in its predictions.
Specifically, we smooth the target distribution with a prior distribution $f$ that is independent of the current input $\source$ \citep{szegedy2015inception,pereyra2017regularize,vaswani2017transformer}.
We use a uniform prior distribution over all words in the vocabulary, \( f = \frac{1}{V} \). One may also use a unigram distribution which has been shown to work better on some tasks \citep{pereyra2017regularize}.
Label smoothing is equivalent to adding the KL divergence between $f$ and the model prediction $p(\candhypo|\source)$ to the negative log likelihood (\TokLS, Equation~\ref{eq:tokls}).
In practice, we implement label smoothing by modifying the ground truth distribution for word $\candhypotok$ to be $q(\candhypotok) = 1 - \epsilon$ and $q(\candhypotok') = \frac{\epsilon}{V}$ for $\candhypotok' \ne \candhypotok$ instead of $q(\candhypotok) = 1$ and $q(\candhypotok') = 0$ where $\epsilon$ is a smoothing parameter.

\subsection{Sequence-Level Objectives}
\label{sec:seqobj}

We also consider a class of objective functions that are computed over entire sequences, i.e., sequence-level objectives.
Training with these objectives requires generating and scoring multiple candidate output sequences for each input sequence during training, which is computationally expensive but allows us to directly optimize task-specific metrics such as BLEU or ROUGE.

Unfortunately, these objectives are also typically defined over the entire space of possible output sequences, which is intractable to enumerate or score with our models.
Instead, we compute our sequence losses over a subset of the output space, $\candhypos(\source)$, generated by the model.
We discuss approaches for generating this subset in \textsection\ref{sec:candidategen}.

\subsubsection*{Sequence Negative Log Likelihood (\SeqNLL)}
\label{sec:seqnll}

Similar to \TokNLL, we can minimize the negative log likelihood of an entire sequence rather than individual tokens (\SeqNLL, Equation~\ref{eq:seqnll}).
The log-likelihood of sequence $\candhypo$ is the sum of individual token log probabilities, normalized by the number of tokens to avoid bias towards shorter sequences:
$$p(\candhypo|\source) = \exp \frac{1}{n}\sum_{i=1}^n \log p(\candhypotok_i|\candhypotok_1,\dots,\candhypotok_{i-1}, \source)$$
As target we choose a pseudo reference\footnote{Another option is to use the gold reference target, $\goldreference$, but in practice this can lead to degenerate solutions in which the model assigns low probabilities to nearly all outputs. This is discussed further in \textsection\ref{sec:candidategen}.
} amongst the candidates which maximizes either
BLEU or ROUGE with respect to $\goldreference$, the gold reference:
\[ \pseudoreference(\source) = \argmax_{\candhypo \in \candhypos(\source)} \operatorname{BLEU}(\goldreference, \candhypo) \]
As is common practice when computing BLEU at the sentence-level, we smooth all initial counts to one (except for unigram counts) so that the geometric mean is not dominated by zero-valued $n$-gram match counts~\citep{lin2004orange}.

\subsubsection*{Expected Risk Minimization (\Risk)}
\label{sec:risk}

This objective minimizes the expected value of a given cost function over the space of candidate sequences (\Risk, Equation~\ref{eq:risk}).
In this work we use task-specific cost functions designed to maximize BLEU or ROUGE~\citep{lin2004rouge}, e.g., $\operatorname{cost}(\goldreference, \candhypo) = 1 - \operatorname{BLEU}(\goldreference, \candhypo)$, for a given a candidate sequence $\candhypo$ and target $\goldreference$.
Different to \SeqNLL~(\textsection\ref{sec:seqnll}), this loss may increase the score of several candidates that have low cost, instead of focusing on a single sequence which may only be marginally better than any alternatives.
Optimizing this loss is a particularly good strategy if the reference is not always reachable,
although compared to classical phrase-based models, this is less of an issue with neural sequence to sequence models that predict individual words or even sub-word units.

The \Risk~objective is similar to the REINFORCE objective used in Ranzato et al.~\shortcite{ranzato2015sequence}, since both objectives optimize an expected cost or reward~\citep{williams1992reinforce}.
However, there are a few important differences:
(1) whereas REINFORCE typically approximates the expectation with a single sampled sequence, the \Risk~objective considers multiple sequences;
(2) whereas REINFORCE relies on a {\it baseline reward}\footnote{Ranzato et al.~\shortcite{ranzato2015sequence} estimate the baseline reward for REINFORCE with a separate linear regressor over the model's current hidden state.} to determine the sign of the gradients for the current sequence, for the \Risk~objective we instead estimate the expected cost over a set of candidate output sequences (see \textsection\ref{sec:candidategen});
and (3) while the baseline reward is different for every word in REINFORCE, the expected cost is the same for every word in risk minimization since it is computed on the sequence level based on the actual cost.

\subsubsection*{Max-Margin}
\label{sec:maxmargin}

\MaxMargin~(Equation~\ref{eq:maxmargin}) is a classical margin loss for structured prediction \citep{mmmn, structure_pred} which enforces a margin between the model scores of the highest scoring candidate sequence $\hat{\candhypo}$ and a reference sequence.
We replace the human reference $\goldreference$ with a pseudo-reference $\pseudoreference$ since this setting performed slightly better in early experiments; $\pseudoreference$ is the candidate sequence with the highest BLEU score.
The size of the margin \emph{varies} between samples and is given by the difference between the cost of $\pseudoreference$ and the cost of $\hat{\candhypo}$.
In practice, we scale the margin by a hyper-parameter $\beta$ determined on the validation set: $\beta(\operatorname{cost}(\goldreference,\hat{\candhypo}) - \operatorname{cost}(\goldreference,\pseudoreference))$.
For this loss we use the unnormalized scores computed by the model before the final softmax:
$$s(\candhypo | \source) = \frac{1}{n} \sum_{i=1}^n s(\candhypotok_{i} | \candhypotok_1, \dots, \candhypotok_{i-1}, \source)$$

\subsubsection*{Multi-Margin}
\label{sec:multimargin}

\MaxMargin~only updates two elements in the candidate set.
We therefore consider \MultiMargin~(Equation~\ref{eq:multimargin}) which enforces a margin between \emph{every} candidate sequence $\candhypo$ and a reference sequence \citep{herbrich199icann}, hence the name Multi-Margin.
Similar to \MaxMargin, we replace the reference $\goldreference$ with the pseudo-reference $\pseudoreference$.

\subsubsection*{Softmax-Margin}
\label{sec:smm}

Finally, \SoftmaxMargin~(Equation~\ref{eq:smm}) is another classic loss that has been proposed by \citet{gimpel+smith2010acl} as another way to optimize task-specific costs.
The loss augments the scores inside the $\exp$ of \SeqNLL~(Equation~\ref{eq:seqnll}) by a cost.
The intuition is that we want to penalize high cost outputs proportional to their cost.

\subsection{Combined Objectives}
\label{sec:comb}

We also experiment with two variants of combining sequence-level objectives (\textsection\ref{sec:seqobj}) with token-level objectives (\textsection\ref{sec:tokobj}).
First, we consider a weighted combination (\Weighted) of both a sequence-level and token-level objective \citep{wu2016google}, e.g., for \TokLS~and \Risk~we have:
\begin{equation}
\mathcal{L}_{\Weighted} = \alpha \mathcal{L}_{\TokLS} + (1 - \alpha) \mathcal{L}_{\Risk}
\label{eq:weighted}
\end{equation}
where $\alpha$ is a scaling constant that is tuned on a held-out validation set.

Second, we consider a constrained combination (\Constrained), where for any given input we use either the token-level or sequence-level loss, but not both.
The motivation is to maintain good token-level accuracy while optimizing on the sequence-level.
In particular, a sample is processed with the sequence loss if the token loss under the current model is at least as good as the token loss of a baseline model $\mathcal{L}^{b}_{\TokLS}$. Otherwise, we update according to the token loss:
\begin{equation}
\mathcal{L}_{\Constrained} =
\begin{cases}
\mathcal{L}_{\Risk} & \mathcal{L}_{\TokLS} \leq \mathcal{L}^{b}_{\TokLS}  \\
\mathcal{L}_{\TokLS} & \text{otherwise}
\end{cases}
\label{eq:constrained}
\end{equation}
In this work we use a fixed baseline model that was trained with a token-level loss to convergence.

\section{Candidate Generation Strategies}
\label{sec:candidategen}

The sequence-level objectives we consider (\textsection\ref{sec:seqobj}) are defined over the entire space of possible output sequences, which is intractable to enumerate or score with our models.
We therefore use a subset of $K$ candidate sequences $\candhypos(\source) = \{\candhypotok_1, \dots, \candhypotok_K\}$, which we generate with our models.

We consider two search strategies for generating the set of candidate sequences.
The first is \emph{beam search}, a greedy breadth-first search that maintains a ``beam" of the top-$K$ scoring candidates at each generation step.
Beam search is the \emph{de facto} decoding strategy for achieving state-of-the-art results in machine translation.
The second strategy is \emph{sampling}~\citep{chatterjee}, which produces $K$ independent output sequences by sampling from the model's conditional distribution.
Whereas beam search focuses on high probability candidates, sampling introduces more diverse candidates (see comparison in \textsection\ref{sec:results_sampling}).

We also consider both online and offline candidate generation settings in \textsection\ref{sec:results_gen}.
In the online setting, we regenerate the candidate set every time we encounter an input sentence $\source$ during training.
In the offline setting, candidates are generated before training and are never regenerated. Offline generation is also embarrassingly parallel because all samples use the same model.
The disadvantage is that the candidates become stale.
Our model may perfectly be able to discriminate between them after only a single update, hindering the ability of the loss to correct eventual search errors.\footnote{We can mitigate this issue by regenerating infrequently, i.e., once every $b$ batches but we leave this to future work.}

Finally, while some past work has added the reference target to the candidate set, i.e., $\candhypos{}'(\source) = \candhypos(\source) \cup \{ \goldreference \}$, we find this can destabilize training since the model learns to assign low probabilities nearly everywhere, ruining the candidates generated by the model, while still assigning a slightly higher score to the reference (cf. \citet{shen2016mrt}).
Accordingly we do not add the reference translation to our candidate sets.

\section{Experimental Setup}
\label{sec:expsetup}

\subsection{Translation}
\label{sec:expsetup_translation}
We experiment on the IWSLT'14 German to English~\citep{cettolo2014report} task using a similar setup as Ranzato et al.~\shortcite{ranzato2015sequence}, which allows us to compare to other recent studies that also adopted this setup, e.g., \citet{wiseman2016acl}.\footnote{Different to \citet{ranzato2015sequence} we train on sentences of up to 175 rather than 50 tokens.}
The training data consists of 160K sentence pairs and the validation set comprises 7K sentences randomly sampled and held-out from the train data.
We test on the concatenation of all available test and dev sets of IWSLT 2014, that is \textit{TED.tst2010, TED.tst2011, TED.tst2012} and \textit{TED.dev2010, TEDX.dev2012} which is of similar size to the validation set.\footnote{In a previous version of this paper, we erroneously quoted the use of \textit{tst2013}. We are using \emph{TEDX.dev2012} instead.}
All data is lowercased and tokenized with a byte-pair encoding (BPE) of 14,000 types \citep{sennrich2016bpe} and we evaluate with case-insensitive BLEU.

We also experiment on the much larger WMT'14 English-French task. 
We remove sentences longer than 175 words as well as pairs with a source/target length ratio exceeding 1.5 resulting in 35.5M sentence-pairs for training.
The source and target vocabulary is based on 40K BPE types.
Results are reported on both newstest2014 and a validation set held-out from the training data comprising 26,658 sentence pairs.

We modify the \emph{fairseq-py} toolkit to implement the objectives described in \textsection\ref{sec:objectives}.\footnote{\url{https://github.com/pytorch/fairseq/tree/classic_seqlevel}.}
Our translation models have four convolutional encoder layers and three convolutional decoder layers with a kernel width of 3 and 256 dimensional hidden states and word embeddings.
We optimize these models using Nesterov's accelerated gradient method \citep{sutskever2013icml} with a learning rate of 0.25 and momentum of 0.99.
  Gradient vectors are renormalized to norm 0.1 \citep{pascanu2013difficulty}.

We train our baseline token-level models for 200 epochs and then anneal the learning by shrinking it by a factor of 10 after each subsequent epoch until the learning rate falls below $10^{-4}$.
All sequence-level models are initialized with parameters of a token-level model before annealing.
We then train sequence-level models for another 10 to 20 epochs depending on the objective.
Our batches contain 8K tokens and we normalize gradients by the number of non-padding tokens per mini-batch.
We use weight normalization for all layers except for lookup tables \citep{salimans2016weight}.
Besides dropout on the embeddings and the decoder output, we also apply dropout to the input of the convolutional blocks at a rate of 0.3~\citep{srivastava2014dropout}.
We tuned the various parameters above and report accuracy on the test set by choosing the best configuration based on the validation set.

We length normalize all scores and probabilities in the sequence-level losses by dividing by the number of tokens in the sequence so that scores are comparable between different lengths.
Additionally, when generating candidate output sequences during training we limit the output sequence length to be less than 200 tokens for efficiency.
We generally use 16 candidate sequences per training example, except for the ablations where we use 5 for faster experimental turnaround.

\subsection{Abstractive Summarization}
\label{sec:expsetup_summary}

For summarization we use the Gigaword corpus as training data \cite{graff2003english} and pre-process it identically to \citet{rush2015abs} resulting in 3.8M training and 190K validation examples.
We evaluate on a Gigaword test set of 2,000 pairs identical to the one used by \citet{rush2015abs} and report F1 ROUGE similar to prior work.
Our results are in terms of three variants of ROUGE \cite{lin2004rouge}, namely, ROUGE-1 (RG-1, unigrams), ROUGE-2 (RG-2, bigrams), and ROUGE-L (RG-L, longest-common substring).
Similar to \citet{ayana2016neural} we use a source and target vocabulary of 30k words.
Our models for this task have 12 layers in the encoder and decoder each with 256 hidden units and kernel width 3.
We train on batches of 8,000 tokens with a learning rate of 0.25 for 20 epochs and then anneal as in \textsection\ref{sec:expsetup_translation}.

\section{Results}
\label{sec:results}

\subsection{Comparison of Sequence Level Losses}
\label{sec:seqlosscmp}

First, we compare all objectives based on a weighted combination with token-level label smoothing (Equation~\ref{eq:weighted}).
We also show the likelihood baseline (MLE) of \citet{wiseman2016acl}, their beam search optimization method (BSO), the actor critic result of \citet{bahdanau2016ac} as well as the best reported result on this dataset to date by \citet{huang2017npbmt}.
We show a like-for-like comparison to \citet{wiseman2016acl} with a similar baseline model below (\textsection\ref{sec:bso}).

Table~\ref{tab:seqresults} shows that all sequence-level losses outperform token-level losses.
Our baseline token-level results are several points above other figures in the literature and we further improve these results by up to 0.61 BLEU with \Risk~training.

\begin{table}[t]
\centering
\begin{tabular}{lrr}
\toprule
& \bf test & \bf std \\ \midrule
MLE (W \& R, 2016) [T] & 24.03 \\
BSO (W \& R, 2016) [S] & 26.36 \\ 
Actor-critic (B, 2016) [S] & 28.53 \\
\citet{huang2017npbmt} [T] & 28.96 \\
\citet{huang2017npbmt} (+LM) [T] & 29.16 \\ 
\midrule
\TokNLL~[T] & 31.78 & 0.07 \\
\TokLS~[T] & 32.23 & 0.10 \\ \midrule
\SeqNLL~[S] & 32.68 & 0.09 \\
\Risk~[S] & 32.84 & 0.08 \\
\MaxMargin~[S] & 32.55 & 0.09  \\
\MultiMargin~[S] & 32.59 & 0.07 \\
\SoftmaxMargin~[S] & 32.71 & 0.07 \\
\bottomrule
\end{tabular}
\caption{Test accuracy in terms of BLEU on IWSLT'14 German-English translation with various loss functions cf. Figure~\ref{fig:objectives}.
W \& R (2016) refers to \citet{wiseman2016acl}, B (2016) to \citet{bahdanau2016ac},
[S] indicates sequence level-training and [T] token-level training. We report averages and standard deviations over five runs with different random initialization. 
}
\label{tab:seqresults}
\end{table}

\subsection{Combination with Token-Level Loss}
\label{sec:results_comb}

Next, we compare various strategies to combine sequence-level and token-level objectives (cf. \textsection\ref{sec:comb}).
For these experiments we use 5 candidate sequences per training example for faster experimental turnaround.
We consider \Risk~as sequence-level loss and label smoothing as token-level loss.
Table~\ref{tab:combresults} shows that combined objectives perform better than pure \Risk.
The weighted combination (Equation~\ref{eq:weighted}) with $\alpha=0.3$ performs best, outperforming constrained combination (Equation~\ref{eq:constrained}).
We also compare to randomly choosing between token-level and sequence-level updates and find it underperforms the more principled constrained strategy.
In the remaining experiments we use the weighted strategy.

\begin{table}[t]
\centering
\begin{tabular}{lrr}
\toprule
& \bf valid & \bf test \\ \midrule
\TokLS & 33.11 & 32.21 \\
\Risk~only & 33.55 & 32.45 \\ \midrule
\Weighted & 33.91 & 32.85 \\
\Constrained & 33.77 & 32.79 \\
Random & 33.70 & 32.61 \\
\bottomrule
\end{tabular}
\caption{Validation and test BLEU for loss combination strategies. We either use token-level
\TokLS~and sequence-level \Risk individually or combine them
as a weighted combination, a constrained combination, a random choice for each sample, cf. \textsection\ref{sec:comb}.
}
\label{tab:combresults}
\end{table}

\subsection{Effect of initialization}
\label{sec:results_init}

So far we initialized sequence-level models with parameters from a token-level model trained with label smoothing. Table~\ref{tab:results_init} shows that initializing weighted \Risk~with token-level label smoothing achieves 0.7-0.8 better BLEU compared to initializing with parameters from token-level likelihood.
The improvement of initializing with \TokNLL~is only 0.3 BLEU with respect to the \TokNLL~baseline, whereas, the improvement from initializing with \TokLS~is 0.6-0.8 BLEU.
We believe that the regularization provided by label smoothing leads to models with less sharp distributions that are a better starting point for sequence-level training.

\begin{table}
\centering
\begin{tabular}{lrr}
\toprule
& \bf valid & \bf test \\ \midrule
\TokNLL & 32.96 & 31.74 \\
\Risk~init with \TokNLL & 33.27 & 32.07 \\
$\Delta$ & +0.31 & +0.33 \\ \midrule
\TokLS & 33.11 & 32.21 \\
\Risk~init with \TokLS & 33.91 & 32.85 \\
$\Delta$ & +0.8 & +0.64 \\
\bottomrule
\end{tabular}
\caption{Effect of initializing sequence-level training (\Risk) with parameters from token-level likelihood (\TokNLL) or label smoothing (\TokLS).
}
\label{tab:results_init}
\end{table}

\subsection{Online vs. Offline Candidate Generation}
\label{sec:results_gen}
Next, we consider the question if refreshing the candidate subset at every training step (online) results in better accuracy compared to generating candidates before training and keeping the set static throughout training (offline).
Table~\ref{tab:onoffline} shows that offline generation gives lower accuracy.
However the online setting is much slower, since regenerating the candidate set requires incremental (left to right) inference with our model which is very slow compared to efficient forward/backward over large batches of pre-generated hypothesis.
In our setting, offline generation has 26 times higher throughput than the online generation setting, despite the high inference speed of fairseq \citep{gehring2017icml}.
\begin{table}
\centering
\begin{tabular}{lrr}
\toprule
& \bf valid & \bf test \\ \midrule
Online generation & 33.91 & 32.85 \\
Offline generation & 33.52 & 32.44 \\
\bottomrule
\end{tabular}
\caption{Generating candidates online or offline.
}
\label{tab:onoffline}
\end{table}

\subsection{Beam Search vs. Sampling and Candidate Set Size}
\label{sec:results_sampling}

\begin{figure}[t]
\includegraphics[width=\linewidth]{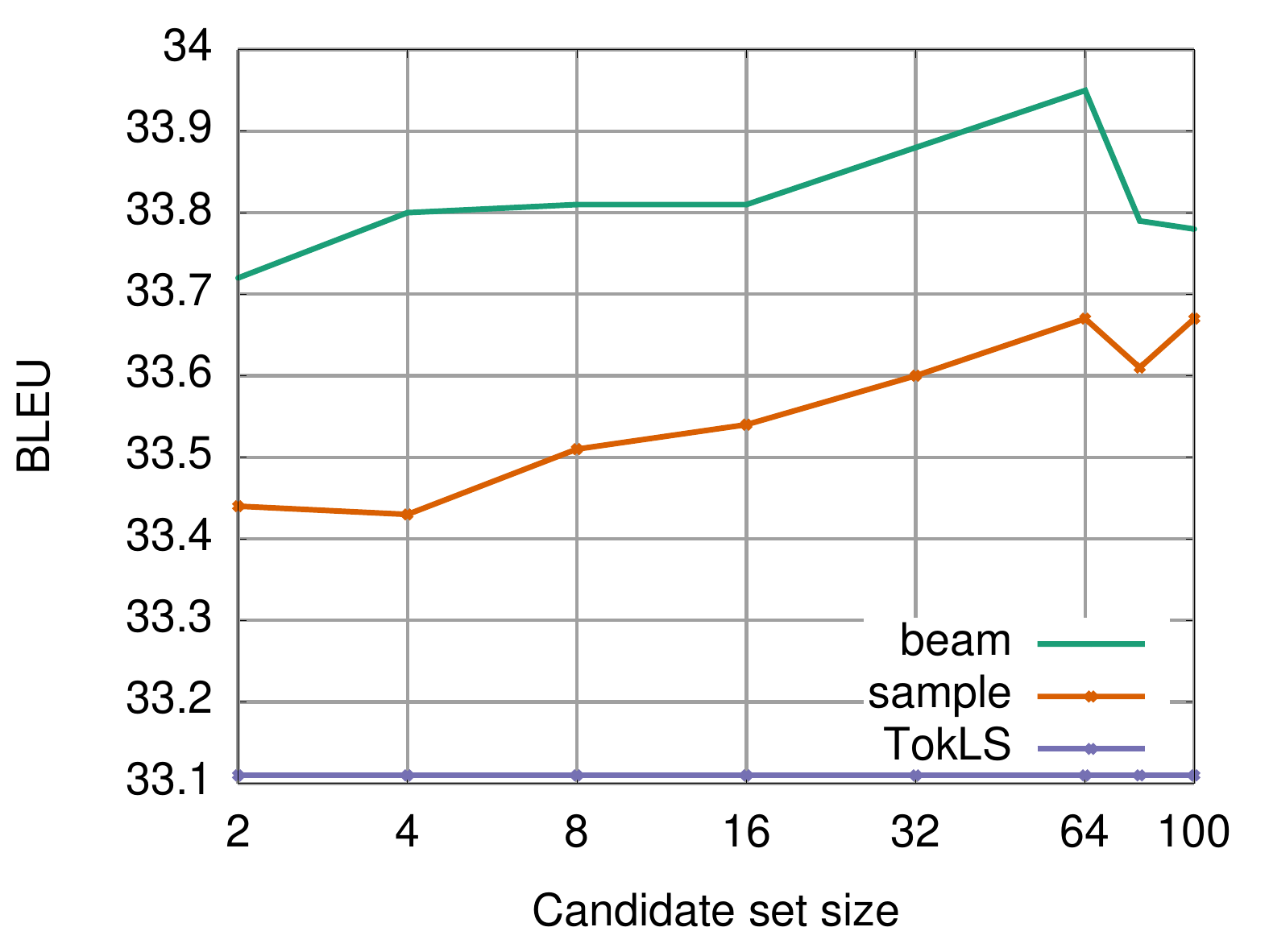}
\caption{Candidate set generation with beam search and sampling for various candidate set sizes during sequence-level training in terms of validation accuracy. Token-level label smoothing (TokLS) is the baseline.}
\label{fig:candset}
\end{figure}

So far we generated candidates with beam search, however, we may also sample to obtain a more diverse set of candidates \citep{shen2016mrt}.
Figure~\ref{fig:candset} compares beam search and sampling for various candidate set sizes on the validation set.
Beam search performs better for all candidate set sizes considered.
In other experiments, we rely on a candidate set size of 16 which strikes a good balance between efficiency and accuracy.

\subsection{Comparison to Beam-Search Optimization}
\label{sec:bso}

\begin{table}[t]
\centering
\begin{tabular}{lrc}
\toprule
& \bf BLEU & $\Delta$ \\ \midrule
MLE & 24.03 & \\
+ BSO & 26.36 & +2.33 \\ 
\midrule
MLE Reimplementation & 23.93 & \\
+ \Risk & 26.68 & +2.75 \\
\bottomrule
\end{tabular}
\caption{Comparison to Beam Search Optimization. We report the best likelihood (MLE) and BSO results from \citet{wiseman2016acl}, as well as results from our MLE reimplementation and training with \Risk.
Results based on unnormalized beam search ($k=5$).
}
\label{tab:bso}
\end{table}

\begin{table}[t]
\centering
\begin{tabular}{lrrr}
\toprule
& \bf RG-1 & \bf RG-2 & \bf RG-L \\ \midrule
ABS+  [T]           & 29.78 & 11.89 & 26.97 \\
RNN MLE [T]    & 32.67 & 15.23 & 30.56 \\
RNN MRT [S]    & 36.54 & 16.59 & 33.44 \\
WFE [T]      & 36.30 & 17.31 & 33.88 \\
SEASS  [T]       & 36.15 & 17.54 & 33.63 \\
DRGD [T]           & 36.27 & 17.57 & 33.62 \\
\midrule
\TokLS                               & 36.53 & 18.10 & 33.93 \\
+ \Risk~RG-1                         & 36.96 & 17.61 & 34.18 \\
+ \Risk~RG-2                         & 36.65 & 18.32 & 34.07 \\
+ \Risk~RG-L                         & 36.70 & 17.88 & 34.29 \\
\bottomrule
\end{tabular}
\caption{Accuracy on Gigaword abstractive summarization in terms of F-measure Rouge-1 (RG-1), Rouge-2 (RG-2), and Rouge-L (RG-L) for token-level label smoothing, and \Risk~optimization of all three ROUGE F1 metrics. [T] indicates a token-level objective and [S] indicates a sequence level objectives. 
ABS+ refers to \citet{rush2015abs}, RNN MLE/MRT \citep{ayana2016neural}, WFE \citep{suzuki2017cutting}, SEASS \citep{zhou2017seass}, DRGD \citep{li2017drgd}.
}
\label{tab:summary}
\end{table}

Next, we compare classical sequence-level training to the recently proposed Beam Search Optimization \citep{wiseman2016acl}.
To enable a fair comparison, we re-implement their baseline, a single layer LSTM encoder/decoder model with 256-dimensional hidden layers and word embeddings as well as attention and input feeding \citep{luong2015effective}.
This baseline is trained with Adagrad~\citep{duchi2011adaptive} using a learning rate of $0.05$ for five epochs, with batches of 64 sequences.
For sequence-level training we initialize weights with the baseline parameters and train with Adam~\citep{kingma2014adam} for another 10 epochs with learning rate $0.00003$ and 16 candidate sequences per training example.
We conduct experiments with \Risk~since it performed best in trial experiments.

Different from other sequence-level experiments (\textsection\ref{sec:expsetup}), we rescale the BLEU scores in each candidate set by the difference between the maximum and minimum scores of each sentence.
This avoids short sentences dominating the sequence updates, since candidate sets for short sentences have a wider range of BLEU scores compared to longer sentences; a similar rescaling was used by \citet{bahdanau2016ac}.

Table~\ref{tab:bso} shows the results from \citet{wiseman2016acl} for their token-level likelihood baseline (MLE), best beam search optimization results (BSO), as well as our reimplemented baseline.
\Risk~significantly improves BLEU compared to our baseline at +2.75 BLEU, which is slightly better than the +2.33 BLEU improvement reported for Beam Search Optimization (cf.~\citet{wiseman2016acl}).
This shows that classical objectives for structured prediction are still very competitive.

\subsection{WMT’14 English-French results}

Next, we experiment on the much larger WMT'14 English-French task using the same model setup as \citet{gehring2017icml}.
We \TokLS for 15 epochs and then switch to sequence-level training for another epoch. 
Table~\ref{tab:wmt} shows that sequence-level training can improve an already very strong model by another +0.37 BLEU. 
Next, we improve the baseline by adding \emph{self-attention} \citep{paulus2017summary,vaswani2017transformer} to the decoder network (\TokLS~+ selfatt) which results in a smaller gain of +0.2 BLEU by \Risk.
If we train \Risk~only on the news-commentary portion of the training data, then we achieve state of the art accuracy on this dataset of 41.5 BLEU \citep{yingce2017deliberation}.

\begin{table}
\centering
\begin{tabular}{lrr}
\toprule
& \bf valid & \bf test \\ \midrule
\TokLS	& 34.06 & 40.58 \\
+ \Risk & 34.20 & 40.95 \\
\midrule
\TokLS~+ selfatt	& 34.24 & 41.02 \\
+ in domain & 34.51 & 41.26 \\
+ \Risk	& 34.30 & 41.22 \\
+ \Risk~in domain & 34.50 & 41.47 \\
\bottomrule
\end{tabular}
\caption{Test and valid BLEU on WMT'14 English-French with and without decoder self-attention. 
}
\label{tab:wmt}
\end{table}

\subsection{Abstractive Summarization}
\label{sec:results_summary}

Our final experiment evaluates sequence-level training on Gigaword headline summarization. 
There has been much prior art on this dataset originally introduced by \citet{rush2015abs} who experiment with a feed-forward network (ABS+).
\citet{ayana2016neural} report a likelihood baseline (RNN MLE) and also experiment with risk training (RNN MRT). Different to their setup we did not find a softmax temperature to be beneficial, and we use beam search instead of sampling to obtain the candidate set (cf. \textsection\ref{sec:results_sampling}). \citet{suzuki2017cutting} improve over an MLE RNN baseline by limiting generation of repeated phrases. \citet{zhou2017seass} also consider an MLE RNN baseline and add an additional gating mechanism for the encoder. \citet{li2017drgd} equip the decoder of a similar network with additional latent variables to accommodate the uncertainty of this task.

Table~\ref{tab:summary} shows that our baseline (\TokLS) outperforms all prior approaches in terms of ROUGE-2 and ROUGE-L and it is on par to the best previous result for ROUGE-1. We optimize all three ROUGE metrics separately and find that \Risk~can further improve our strong baseline.
We also compared \Risk~only training to \Weighted~on this dataset (cf. \textsection\ref{sec:results_comb}) but accuracy was generally lower on the validation set: RG-1 (36.59 \Risk~only vs. 36.67 \Weighted), RG-2 (17.34 vs. 18.05), and RG-L (33.66 vs. 33.98).

\section{Conclusion}

We present a comprehensive comparison of classical losses for structured prediction and apply them to a strong neural sequence to sequence model.
We found that combining sequence-level and token-level losses is necessary to perform best, and so is training on candidates decoded with the current model.

We show that sequence-level training improves state-of-the-art baselines both for IWSLT'14 German-English translation and Gigaword abstractive sentence summarization. Structured prediction losses are very competitive to recent work on reinforcement or beam optimization. 
Classical expected risk can slightly outperform beam search optimization \citep{wiseman2016acl} in a like-for-like setup.
Future work may investigate better use of already generated candidates since invoking generation for each batch slows down training by a large factor, e.g., mixing with fresh and older candidates inspired by MERT~\cite{och:2003:acl}.

\bibliography{master}
\bibliographystyle{acl_natbib}

\end{document}